%% file: main.tex
\documentclass{l4dc2025}
\usepackage{dsfont}
\usepackage{amsmath}
\usepackage{booktabs}
\usepackage{bbm}
\usepackage{wrapfig}
\usepackage{multirow}
\usepackage{enumitem}

\SetKwInput{KwInput}{Input}                
\SetKwInput{KwOutput}{Output}              

\usepackage{algorithm,algcompatible}
\algnewcommand\INPUT{\item[\textbf{Input:}]}%
\algnewcommand\OUTPUT{\item[\textbf{Output:}]}%

\usepackage{subcaption}
\usepackage{url}
\usepackage{wrapfig}

\DeclareMathOperator*{\argmax}{arg\,max}
\DeclareMathOperator*{\argmin}{arg\,min}
\newcommand{\bestresponse}{\mathcal{BR}}

\newtheorem{mydef}{Definition}

\newtheorem{myprob}{Problem}

\title[STL Game]{STLGame: Signal Temporal Logic Games \\in Adversarial Multi-Agent Systems}
\usepackage{times}

\author{%
 \Name{Shuo Yang}$^{1}$ \Email{yangs1@seas.upenn.edu}
 \AND
 \Name{Hongrui Zheng}$^{1}$ \Email{hongruiz@seas.upenn.edu}
 \AND
 \Name{Cristian-Ioan Vasile}$^{2}$ \Email{cvasile@lehigh.edu}
 \AND
 \Name{George J. Pappas}$^{1}$ 
\Email{pappasg@seas.upenn.edu}
\AND
 \Name{Rahul Mangharam}$^{1}$ \Email{rahulm@seas.upenn.edu}\\
 \addr $^{1}$University of Pennsylvania, Philadelphia, PA, USA \\
 \addr $^{2}$Lehigh University, Bethlehem, PA, USA \\
}

\begin{document}

\maketitle

\begin{abstract}%
We study how to synthesize a robust and safe policy for autonomous systems under signal temporal logic (STL) tasks in adversarial settings against unknown dynamic agents. To ensure the worst-case STL satisfaction, we propose STLGame, a framework that models the multi-agent system as a two-player zero-sum game, where the ego agents try to maximize the STL satisfaction and other agents minimize it. STLGame aims to find a Nash equilibrium policy profile, which is the best case in terms of robustness against unseen opponent policies, by using the fictitious self-play (FSP) framework. FSP iteratively converges to a Nash profile, even in games set in continuous state-action spaces. We propose a gradient-based method with differentiable STL formulas, which is crucial in continuous settings to approximate the best responses at each iteration of FSP. We show this key aspect experimentally by comparing with reinforcement learning-based methods to find the best response. Experiments on two standard dynamical system benchmarks, Ackermann steering vehicles and autonomous drones, demonstrate that our converged policy is almost unexploitable and robust to various unseen opponents' policies.
All code and additional experimental results can be found on our project website: \url{https://sites.google.com/view/stlgame}
\end{abstract}

\begin{keywords}%
  Signal temporal logic, game theory, multi-agent systems%
\end{keywords}

\input{intro}
\input{preliminary}
\input{formulation}
\input{method}
\input{simulation}
\input{conclude}


\bibliography{reference}

\input{append}

\end{document}

%% file: intro.tex
\section{Introduction}


Safety-critical autonomous systems operate in dynamic environments to facilitate human life.
These autonomous systems usually have complex tasks that are critical for safety and time and have logical dependency between tasks.
For instance, Amazon drones\footnote{\url
{https://www.aboutamazon.com/news/transportation/amazon-drone-delivery-arizona}} may need to deliver items to customers within a specified time and visit a particular warehouse every hour; in addition, safety, such as no collision with other drones or obstacles, is the top priority.
These complex task and safety specifications can be expressed naturally as a temporal logic formula~\cite{fainekos2009temporal, smith2011optimal}, which combines temporal and logical operators to formally define these tasks in a rigorous way. Dynamic environments typically involve multiple active agents.
Multi-agent systems control under temporal logic specifications has been studied in the past mainly assuming agents are cooperative~\cite{lindemann2019control, sun2022multi}.
However, this does not hold in many cases.
Other agents can be independent or even adversarial to ego agents.
For instance, drones from different companies can be sharing the same airspace, posing risks on the task completion and even safety of each other.
Moreover, other agents may adopt non-stationary policies, which makes it even more challenging to synthesize a safe and robust policy.

In this work, we consider the problem where the ego agents'\footnote{Hereafter, we use \emph{ego agents} to denote those controllable agents, and use \emph{other agents} and \emph{opponents} interchangeably to denote those uncontrollable agents.} task is specified by a signal temporal logic (STL) formula~\cite{maler2004monitoring}, and we desire to synthesize a control policy for ego agents such that the STL task can be satisfied.
The challenging aspect here is that the ego agents share the environment with other a-priori unknown, independent, and potentially adversarial agents.
To ensure that STL task is satisfied as much as possible even in the worst case, we formulate \textbf{STLGame} (Signal Temporal Logic Game) by modeling the system as a two-player zero-sum stochastic game where ego agents try to maximize STL satisfaction and opponents aim to minimize the same STL satisfaction.
STLGame finally concerns finding a Nash Equilibrium (NE) policy for ego agents.
The advantage of using an NE policy is that it provides the best-case expected returns against any possible unknown opponent agents.
STLGame may be somewhat conservative in practice because other agents are not necessarily adversarial, but it provides the safest and most robust solution when no information about other agents is revealed beforehand.

The proposed STLGame essentially asks us to find a Nash equilibrium policy profile in continuous environments where the level of STL satisfaction/robustness is the utility function.
However, converging to Nash equilibrium is generally extremely challenging, if not impossible, in continuous action spaces.
We leverage the classic Fictitious Play (FP)~\cite{brown1951iterative} framework from the game theory community in this work, which provably converges to NE in two-player zero-sum games. In addition, another key aspect of enabling STLGame's success in the continuous space is leveraging differentiable STL formulas~\cite{pant2017smooth, leung2023backpropagation}, which significantly improves sample efficiency when optimizing a control policy. 
To summarize, our contributions are as follows:

\begin{enumerate}[topsep=1pt,itemsep=-1ex]
    \item We formulate STLGame, a robust STL synthesis framework that ensures worst-case STL satisfaction against dynamic opponent agents, and use the iterative fictitious play to converge to Nash equilibrium policies;
    \item To compute the best response policy at each iteration of FP, we propose an STL gradient-based method that can synthesize an approximate best response against opponents' policies;
    \item We compare the gradient-based method with model-free reinforcement learning for best responses and empirically demonstrate its effectiveness, convergence and sample efficiency;
    \item We implement our framework on Ackermann steering vehicles and autonomous drones, and empirically show that the converged policy is almost unexploitable, and robust to unseen opponents.
    Our code is publicly available\footnote{\url{https://github.com/shuoyang2000/STLgame}}.
\end{enumerate}
\vspace{-4mm}
\subsection{Related Work}
\noindent \textbf{Temporal Logic Games}
With (linear) temporal logic (LTL) specifications, games are usually studied in abstracted transition systems with finite discrete states, such as B\"{u}chi game and Rabin game~\cite{baier2008principles, belta2017formal}, where the non-determinism in the environment is modeled as the second player~\cite{kloetzer2012ltl, cui2023towards}.
Our work instead considers games with STL specification which is evaluated on real-time and real-valued signals, and the system has continuous state-action spaces, so the game complexity is much higher.

\cite{vande2023game} formulates a potential game for multi-agent systems under STL specifications, assuming individual agent’s utility function is aligned with the global objective function.
\cite{yu2023signal} also considers multi-agents control under STL tasks where opponents' policy is unknown, but assuming opponents' policy is stationary and some data reflecting the policy can be accessed.
Unlike these two works,
we consider the adversarial case and do not assume stationary policy or policy data access on opponents.
The closest work to ours is~\cite{muniraj2018enforcing}, in which the authors also consider STL zero-sum games and use minimax Q-learning~\cite{littman1994markov} to converge to Nash policy.
However, \cite{muniraj2018enforcing} mainly focuses on discrete action spaces as it is not clear how to effectively apply minimax Q-learning to continuous spaces without significantly sacrificing control performance.
Our work instead bridges this gap using fictitious play and differentiable STL formulas. 
To the best of our knowledge, this is the first work to study temporal logic-based adversarial games in continuous state-action spaces.

\noindent \textbf{Fictitious Play} For many discrete games such as poker, counterfactual regret minimization (CFR) method and its variants~\cite{zinkevich2007regret, brown2019superhuman} are state-of-the-art.
In general, it is still not clear how to extend CFR into continuous games, although there is some domiain-specific results~\cite
{zheng2024bridging}.
Fictitious self-play~\cite{heinrich2015fictitious}, on the other hand, can generalize to continuous games easily~\cite{goldstein2022converging} where a best response policy in continuous space can be computed at each iteration.

%% file: preliminary.tex
\vspace{-10pt}

\section{Background}
\setlength{\belowdisplayskip}{2pt} \setlength{\belowdisplayshortskip}{2pt}
\setlength{\abovedisplayskip}{2pt} \setlength{\abovedisplayshortskip}{2pt}
\subsection{Signal Temporal Logic}
In this work, we use signal temporal logic (STL) to describe the tasks that ego agents aim to achieve.
STL was introduced in \cite{maler2004monitoring} and its syntax is defined as
\begin{align}\label{eq:stl}
    \phi ::= \texttt{True}\mid\pi^\mu \mid \neg \phi \mid \phi_1 \wedge \phi_2 \mid \phi_1 \textbf{U}_{[a,b]} \phi_2,
\end{align}
where $\pi^\mu:\mathbb{R}^n \to \{\texttt{True}, \texttt{False} \}$ is a predicate whose truth value   is determined by the sign of the function $\mu:\mathbb{R}^n \to \mathbb{R}$.
The symbols $\neg$ and $\wedge$ denote the Boolean operators ``negation" and ``conjunction" respectively, and $a, b$ are non-negative
scalars ($a\le b$) used to denote time lower and upper bounds respectively. $\textbf{U}_{[a,b]}$ denotes the temporal operator ``\emph{until}".
Note that these basic operators can be combined to define some other commonly-used temporal and logical operators such as
``eventually" $\Diamond_{[a,b]} \phi:= \texttt{True} \mathbf{U}_{[a,b]} \phi$,
``always" $\Box_{[a,b]} \phi:=\neg \Diamond_{[a,b]} \neg \phi$,
``disjunction" $\phi_1 \vee \phi_2:=\neg(\neg \phi_1 \wedge \neg \phi_2)$,
and ``implication" $\phi_1 \to \phi_2:= \neg \phi_1 \vee   \phi_2$.

\noindent\textbf{STL Qualitative Semantics.} STL specifications are evaluated over signal/trajectory 
$\mathbf{s} =
 s_0s_1\cdots$.  
We use $(\mathbf{s},k) \models \phi$ to denote that $\mathbf{s}$ satisfies the STL formula $\phi$ at time $k$. Formally, STL semantics are defined as follows:
\begin{equation}
\label{eq:stl-semantics}
  \begin{aligned}
    (\mathbf{s}, k) \models \pi^{\mu} &\iff  \mu(s_k) \geq 0, \quad (\mathbf{s},k)\models \neg \phi\iff (\mathbf{s},k)\not\models \phi,\\
    (\mathbf{s},k)\models \phi_1 \wedge \phi_2&\iff (\mathbf{s},k)\models \phi_1 \text{ and } (\mathbf{s},k)\models \phi_2,\\
(\mathbf{s},k)\models \phi_1 \mathbf{U}_{[a,b]} \phi_2 &\iff \exists k' \!\in\! [k+a, k+b], \text{s.t. } (\mathbf{s},k') \models \phi_2.
\text{ and }(\mathbf{s},k'') \models \phi_1, \forall k'' \!\in\! [k, k'].\nonumber
  \end{aligned}  
\end{equation}
The intuition of $\phi_1 \mathbf{U}_{[a,b]} \phi_2$ is that $\phi_2$ can hold some time between $[k + a, k + b]$ in the future
and $\phi_1$ will always hold until then.
For convenience, we write $\mathbf{s} \models \phi$ instead of $(\mathbf{s},0) \models \phi$.

\noindent\textbf{STL Quantitative Semantics.}
In addition to the STL qualitative Boolean semantics, we can also define the STL \emph{quantitative} semantics $\rho^\phi(\mathbf{s},k)\in\mathbb{R}$, also called \emph{robustness}, that indicates how much $\phi$ is satisfied or violated \cite{fainekos2009robustness, donze2010robust}.
The robustness degree $\rho^{\phi}(\mathbf{s}, k)$ is defined as follows:
\begin{equation}
\label{eq:stl-robustness}
\begin{aligned}
\rho^{\texttt{True}}(\mathbf{s}, k) & = \infty, ~~~~~~~~~~~~~~~~~~~~~~~ \rho^{\pi^{\mu}}(\mathbf{s}, k) = \mu(\mathbf{s}_k),\\
    \rho^{\lnot \phi}(\mathbf{s}, k) & = -\rho^{\phi}(\mathbf{s}, k), ~~~~~~ 
    \rho^{\phi_1 \land \phi_2}(\mathbf{s}, k) {}= \min(\rho^{\phi_1}(\mathbf{s}, k), \rho^{\phi_2}(\mathbf{s}, k)),\\
    \rho^{\phi_1 \mathbf{U}_{[a,b]} \phi_2}(\mathbf{s}, k) & = \max_{k'\in [k+a, k+b]} \min(\rho^{\phi_2}(\mathbf{s}, k'), \min_{k''\in[k, k']}\rho^{\phi_1}(\mathbf{s}, k'')).\nonumber
\end{aligned}
\end{equation}
For any trajectory $\mathbf{s}$ and STL $\phi$, we have that $\mathbf{s}\models \phi$ if and only if $\rho^{\phi}(\mathbf{s}, 0)\ge 0$.
Similarly, we write $\rho^{\phi}(\mathbf{s})$ instead of $\rho^{\phi}(\mathbf{s}, 0)$ for simplicity.
In addition, we define the horizon length $T_{\phi}$ of an STL $\phi$ as the duration of time needed to verify whether a signal satisfies $\phi$~\cite{dokhanchi2014line, sadraddini2015robust}. 
For instance, the horizon length of $\phi=\Diamond_{[1,10]} \Box_{[2,5]} \pi^{\mu}$ is $15$.

\vspace{-5pt}

\subsection{Stochastic Game}
We consider the stochastic game~\cite{shapley1953stochastic}
$G=(\mathcal{N}, S, A, f, r, \rho_0, T, \gamma)$ with $N$ agents, 
where $\mathcal{N}=\{1, 2, \cdots, N\}$ denotes the set of $N$ agents,
$S:=\times_{i \in \mathcal{N}}~S^i\subseteq \mathbb{R}^n$ and $A:=\times_{i \in \mathcal{N}}~A^i$ are the set of states and joint actions respectively,
$f : S\times A \rightarrow S$ is the deterministic transition function,
$r^i : S \times A \times S \rightarrow \mathbb{R}$ is the reward function for agent $i$,
$\rho_0$ represents the distribution of initial conditions,
$T \in \mathbb{N}$ denotes the time horizon,
$\gamma \in [0, 1]$ is the discount factor.
At timestep $t$, 
each agent $i$ picks an action $a^i_t$ according to its (potentially stochastic) policy $\pi^i\in \Pi^i: \sum_{a^i\in A^i}\pi^i(a^i|s)=1$ and 
the system state evolves according to the joint action $a_t = \times_{i \in \mathcal{N}}~a^i_t$ by discrete-time dynamics
\begin{equation}
s_{t+1} = f(s_t, a_t),
\label{eq:transition_dynamics}
\end{equation}
where $f$ is a continuous and nonlinear function in $S\times A$.
A joint policy  is $\pi = (\pi^1, \pi^2, \cdots, \pi^N)$.

\begin{mydef}
A \emph{partially observable
stochastic game (POSG)}~\cite{hansen2004dynamic} is defined by the same elements of a stochastic game and additionally defines for each agent $i\in\mathcal{N}$: 1) a set of observations $O_i$, and 2) observation function $\mathcal{O}_i: A\times S\times O_i\rightarrow [0, 1]$.
\end{mydef}
Following~\cite{marl-book}, in a POSG, we define the \emph{history} by 
$
\hat{h}_t=\{(s_{\tau}, o_{\tau}, a_{\tau})_{\tau=0}^{t-1}, s^t, o^t\}
$
up to time $t$, consisting of the joint states, observations, and actions of all agents in each time step before $t$, and
the current state $s_t$ and current observation $o_t$.
We use $\sigma(\hat{h})$ to denote the history of observations from the full history $\hat{h}$.
We denote $s(\hat{h})=s_t$ the last state in $\hat{h}$ and denote $\langle \rangle$ the concatenation operator.
For any joint policy profile $\pi=\langle \pi^{i}, \pi^{-i} \rangle$ where $-i$ denotes all other agents except $i$ and any history $\hat{h}$, two interlocked functions $V$ and $Q$ are defined as
\begin{align}
V^i_\pi(\hat{h})&=\sum_{a \in A} \pi(a \mid \sigma(\hat{h})) Q^i_\pi(\hat{h}, a),\\
    Q^i_\pi(\hat{h}, a)&=r^i(s(\hat{h}), a, s')+\gamma \sum_{o'\in O}\mathcal{O}(o'|a, s')V^i_\pi(\langle\hat{h}, a, s', o'\rangle), \text{where } s'=f(s(\hat{h}), a),
\end{align}
where $V^i_\pi(\hat{h})$ is the expected return (or value function) for agent $i$ at the history $\hat{h}$ under the joint policy profile $\pi$, 
and $Q^i_\pi(\hat{h}, a)$ is the expected return (or $Q$ function) for agent $i$ when it executes the action $a$ after the history $\hat{h}$ under the joint policy profile $\pi$.
Note that the policy chooses actions based on observation history here instead of the current state like in vanilla stochastic games.
We define the \emph{expected return} for agent $i$ from the initial state of the POSG as
$U^i(\pi)=\mathbb{E}_{s_0 \sim \rho_0, o_0 \sim \mathcal{O}\left(\cdot \mid \emptyset, s_0\right)}[V^i_{\pi}(s_0, o_0)].
$
The set of \emph{best response} (BR) policies for agent $i$ against $\pi^{-i}$ is defined as
\[
\bestresponse(\pi^{-i}):=\argmax_{\pi^i\in\Pi^i} U^i(\langle\pi^i, \pi^{-i}\rangle),
\]
where $\langle\pi^i, \pi^{-i}\rangle$ is the joint policy profile consisting of $\pi^i$ and $\pi^{-i}$.
We write $U^i(\pi^i, \pi^{-i})$ instead of $U^i(\langle\pi^i, \pi^{-i}\rangle)$ for simplicity.
Note that there may exist many BR policies w.r.t. to a $\pi^{-i}$.
The set of \emph{$\epsilon$-approximate best response} (BR) policies for agent $i$ against $\pi^{-i}$ is defined as
\[
\bestresponse_{\epsilon}(\pi^{-i}):=\{\pi^i\in \Pi^i:U^i(\pi^i, \pi^{-i})\ge U^i(\bestresponse(\pi^{-i}), \pi^{-i})-\epsilon\}.
\]
A \emph{Nash equilibrium}  is a joint policy profile $\pi$ such that $\pi^{i}\in \bestresponse(\pi^{-i})$ for any $i\in\mathcal{N}$. 
For $\epsilon>0$, an $\epsilon$-Nash equilibrium is a joint policy profile $\pi$ such that $\pi^{i}\in \bestresponse_{\epsilon}(\pi^{-i})$ for any $i\in\mathcal{N}$.
We then define the \emph{exploitability} of any joint policy profile $\pi$ as
\begin{align}
\label{eq:exploitability}
    \mathcal{E}(\pi):=  \sum_{i\in\mathcal{N}} \Bigl( U^i(\bestresponse(\pi^{-i}), \pi^{-i})  - U^i(\pi^i, \pi^{-i})\Bigr).
\end{align}
Intuitively, exploitability can measure the distance between the current policy and a Nash equilibrium policy.
Note that when the exploitability $\mathcal{E}(\pi)$ is 0, then $\pi$ is a Nash equilibrium policy profile.
And exploitability of $\epsilon$ yields at least an $\epsilon$-Nash equilibrium.
In addition, a stochastic game
is a \emph{zero-sum game} if $\sum_{i\in\mathcal{N}}r^i(s, a, s')=0$ for any state-action pair $(s, a)$ and next state $s'=f(s, a)$.

%% file: formulation.tex
\vspace{-8pt}

\section{Problem Formulation}
In this work, we consider the two-player zero-sum stochastic game.
In other words, we have two teams of agents with one team of $M$ agents (ego agents) trying to maximize a utility function and another team of $N-M$ agents (opponent agents) trying to minimize the utility function.
We assume that all agents within a team are fully cooperative.
Without loss of generality, we assume that $N=2$ and $M=1$ in this work, i.e., each team has only one agent.\footnote{The results developed in this work can be transferred to the case with more agents in each team. However, more agents lead to higher complexity to learn the cooperative policy inside the team.
We leave it for future work.}

The mission that the ego agent is expected to achieve is expressed by an STL formula $\phi$ with bounded horizon $T_{\phi}=T>0$.
Since the quantitative satisfaction of STL is evaluated on the full trajectory $\mathbf{s}=s_0s_1\cdots s_T$, the reward function for ego agent under a general STL specification $\phi$ is formally defined as
\begin{equation}
\label{eq:reward_func_robustness}
  r^1(s(\hat{h}_t), a, s')=
    \begin{cases}
      0 & \text{if $t<T$}\\
      \rho^{\phi}(\mathbf{s}) & \text{if $t=T$}
    \end{cases}       
\end{equation}
and the reward function for opponent agent is defined as $r^{-1}(s(\hat{h}_t), a, s') = - r^1(s(\hat{h}_t), a, s')$.
One can define less sparser rewards as in~\cite{aksaray2016q} for special specifications where $\phi=\Diamond_{[0, T]}\varphi$ or $\phi=\Box_{[0, T]}\varphi$.
The main problem considered in this paper is formulated as follows.
\begin{myprob}
Given the POSG $G$ with dynamical system~(\ref{eq:transition_dynamics}), given an STL formula $\phi$, 
synthesize policies $\pi^{1, *}$ and $\pi^{-1, *}$ for ego and opponent agents respectively such that:
\begin{align}\label{eq:minimax}
    \pi^{1, *} = \argmax_{\pi^1 \in \Pi^1} \min_{\pi^{-1} \in \Pi^{-1}} \mathop{\mathbb{E}}_{\mathbf{s}\sim \pi} \rho^{\phi}(\mathbf{s}),\quad\!\!\!
    \pi^{-1, *} = \argmin_{\pi^{-1} \in \Pi^{-1}} \max_{\pi^{1} \in \Pi^{1}} \mathop{\mathbb{E}}_{\mathbf{s}\sim \pi} \rho^{\phi}(\mathbf{s}),
\end{align}
where $\mathbf{s}\sim \pi$ means that the trajectory $\mathbf{s}$ is sampled under policy profile $\pi$.
\vspace{-5pt}
\end{myprob}
In other words, the ego (opponent) agent is synthesizing a policy that can maximize (minimize) the STL satisfaction robustness under the worst case from the opponent (ego).
It can also be interpreted as a robust policy that optimizes the worst-case return, and it is essentially the best choice when the agent is unaware of the uncertainty in the environment or what policy the other agent will use.
Any policy solution profile $\pi^{*}=(\pi^{1, *}, \pi^{-1, *})$ from~(\ref{eq:minimax}) is called a \emph{minimax solution}.
Interestingly, in two-player zero-sum games, the set of minimax solutions coincides with the set of Nash equilibrium~\cite{owen2013game}.
Therefore, to solve~(\ref{eq:minimax}), we need to find a Nash policy profile.

%% file: method.tex
\vspace{-5pt}

\section{Fictitious Play for STLGame}

Fictitious Play~\cite{brown1951iterative} is an iterative algorithm that can provably converges to a Nash equilibrium in two-player zero-sum games.
The players (ego and opponent agents) play
the game repeatedly, and each player adopts a best response policy to the average policy of the other player at each iteration.
As shown in Algorithm~(\ref{algorithm:fp}), the agents initialize their average policy using a random policy.
At each iteration $k$,
the agent computes a best response policy $\beta^i_{k+1}$ to the other agent’s current average policy $\pi_k^{-i}$
and then update its average policy by 
\begin{align}
\label{eq:fp_avg}
    \pi^{i}_{k+1} = \frac{k}{k+1}\pi^{i}_{k}+\frac{1}{k+1}\beta^{i}_{k+1}.
\end{align}
\begin{wrapfigure}{H}{0.4\textwidth}
\vspace{-55pt}
\begin{minipage}
{0.4\textwidth}
    \begin{algorithm}[H]
\SetAlgoNoEnd\caption{Fictitious Play}
\label{algorithm:fp}
    Initialize $\pi^1_0, \pi^{-1}_0$ with random policies\;
    
    \For{$k=0,\ldots,K$}
    { 
    \For{$i=1, -1$}
    {$\beta^{i}_{k+1} = \bestresponse(\pi^{-i}_k)$\;

    $\pi^{i}_{k+1} = \frac{k}{k+1}\pi^{i}_{k}+\frac{1}{k+1}\beta^{i}_{k+1}$\;
    }  
    }
\end{algorithm}
  \end{minipage}
\vspace{-10pt}
\end{wrapfigure}
The key challenge in fictitious play is the computation of best response policy, especially in environments with continuous action space.
To mitigate this issue, a more general version of fictitious play, called \emph{generalized weakened fictitious play}~\cite{leslie2006generalised}, is proposed by tolerating approximate best responses and perturbed average policy while still converging to Nash equilibrium.
\begin{mydef}
A generalized weakened fictitious play is a process of mixed strategies $\{\pi_k\in \Pi\}$:
$$\pi_{k+1}^i \in\left(1-\alpha_{k+1}\right) \pi_k^i+\alpha_{k+1}\left(\bestresponse_{\epsilon_k}(\pi_k^{-i})+M_{k+1}^i\right), \forall i\in \{1, -1\},$$
with $\alpha_k\rightarrow 0$ and $\epsilon_k\rightarrow 0$ when $k\rightarrow \infty$, $\sum_{k=1}^{\infty}\alpha_k=\infty$, and
$M_k$ is a sequence of perturbations that satisfies
\vspace{-2mm}
$$\lim _{k \rightarrow \infty} \sup _j\left\{\left\|\sum_{l=k}^{j-1} \alpha_{l+1} M_{l+1}\right\|\text{s.t. }\sum_{l=k}^{j-1} \alpha_{l+1} \leq J\right\}=0, \forall J\ge 0.$$
\end{mydef}
One can notice that naive fictitious play is a special case of generalized weakened fictitious play with $\alpha_k=\frac{1}{k}$ (average step size), $\epsilon_k=0$ (strict best response) and $M_k=0$ (no perturbation).

\subsection{Best Responses in STLGame}
The main step of using the (generalized weakened) fictitious play is still the computation of (approximate) best responses.
In the context of STLGames, the best response computation at each iteration is synthesizing the optimal control policy given an opponent agent adopting a fixed but unknown average policy (i.e., $\pi^{-1}_k$ at iteration $k$) in the environment.
Notice that the opponent policy $\pi^{-1}_{k}$ is fixed, stochastic, and unknown to the ego agent,
so the best response computation for the ego agent in STLGame is formulated as the following problem\footnote{Opponent agent's BR synthesis problem can be formulated in a dual manner.
We only analyze the case of an ego agent here for simplicity.}:

\begin{myprob}\label{problem:br}
    Given the POSG $G$ with dynamical system~(\ref{eq:transition_dynamics}) where the opponent agent is playing an unknown policy $\pi^{-1}_k$, given an STL formula $\phi$, synthesize an optimal control policy that maximizes STL robustness: $\pi^{1, *}_k = \argmax_{\pi^1_k \in \Pi^1} \mathop{\mathbb{E}}_{\mathbf{s}\sim \langle\pi^1_k, \pi^{-1}_k\rangle} \rho^{\phi}(\mathbf{s})$.
\end{myprob}
Due to the fact that the opponent has an unknown policy, it is not clear how to use classic approaches such as mixed-integer linear program (MILP)-based method~\cite{milp_stl} to synthesize an optimal policy, as they typically assume access to the precise model of the entire environment.
Regardless of the lack of knowledge on opponent's policy, we can still interact with it and gradually have better understanding on its policy,
and then compute/learn a (near)-optimal policy.
This motivates us to use reinforcement learning to (approximately) solve Problem~(\ref{problem:br}).

\noindent\textbf{Reinforcement Learning for BR.}
Some existing work tried to leverage reinforcement learning (RL) techniques to facilitate STL policy synthesis in both single agent system~\cite{li2017reinforcement, venkataraman2020tractable, wang2024tractable} and multi agents system~\cite{wang2023multi}.
RL training is guided by a reward function, which is the robustness value of the rollout trajectory w.r.t. the STL specification in our case.
For a general STL formula $\phi$, the reward function is defined as~(\ref{eq:reward_func_robustness}).
One can use any model-free RL methods such as Q-learning~\cite{watkins1992q} for environments with discrete action space and Proximal Policy Optimization (PPO)~\cite{schulman2017proximal} for continuous environments.

However, the reward function in~(\ref{eq:reward_func_robustness}) is sparse because the robustness value is not accessible until the end of the trajectory for a general specification.
Learning with sparse reward signals is a challenging exploration problem for RL algorithms, especially for continuous environments.
The sample efficiency for sparse reward setting may be concerning, which influences the (approximate) best response quality if the exploration is terminated once the computation budget is depleted.
Finally, the iterated fictitious play process may be unstable due to the under-par response policy.

\begin{wrapfigure}{r}
{0.46\textwidth}
\vspace{-20pt}
  \begin{center}    \includegraphics[width=0.46\textwidth]{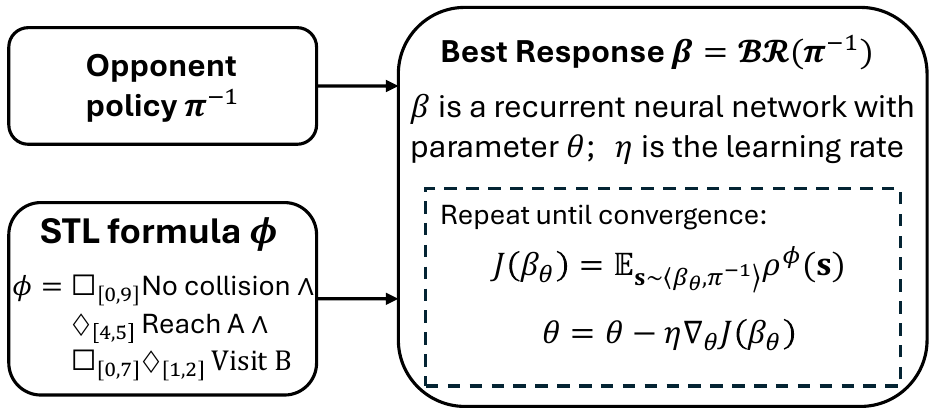}
  \end{center}
  \vspace{-15pt}
  \caption{STL Gradient-based Best Response}
  \label{fig:stl_grad_br}
\end{wrapfigure}

\noindent\textbf{Gradient-based Method for BR.} One drawback of using reinforcement learning for BR synthesis is that it is unaware of the STL information but only of the STL final robustness values.
However, STL has rich information such as its gradients, has not been used by RL when computing a response policy.
Recent work~\cite{pant2017smooth, leung2023backpropagation} shows that STL robustness formulas can be represented as computation graphs, enabling robustness gradient computation with respect to the action.
Thanks to the STL robustness gradients, we can directly use gradient descent algorithm to learn a control policy represented by a neural network (see Figure~(\ref{fig:stl_grad_br})).
Specifically, at each training iteration, we rollout the system with the policy network, compute the STL robustness value over full trajectories, and update the policy network parameters to maximize robustness via backpropagation and gradient descent.
This gradient-based method is expected to require significantly less computation budget than using RL. Thus, the trained policies under the same computation budget are expected to be much closer to an actual best response.

\vspace{-6mm}
We implement and compare both the reinforcement learning and the gradient-based method to compute the best responses in the experiments.
To calculate the average policy in fictitious play, one can represent the average policy using a single stochastic network, and train this network using a supervised learning approach to mimic the behavior of the ground truth average policy; see e.g., \cite{heinrich2015fictitious}.
In the experiment, we observe that using the supervised learning-based average policy does not work well in STL games, so we follow the average rule~(\ref{eq:fp_avg}) directly by sampling the set of found BR policies according to the sampling weight produced by (\ref{eq:fp_avg}) and it produces more stable and less exploitable results.

\noindent \textbf{BR Policy Representation} 
Unlike traditional RL/control methods, which make decisions mainly based on the current observation or state, STL control policy need to also depend on the history observations/states sequence for two reasons:
1) STL formula is evaluated over the whole trajectory so the future trajectory should highly depend on the past trajectory;
2) ego/opponent agent's policy information is implicitly encoded into the history observation sequence and single observation does not reflect the policy.
Therefore, we use Recurrent Neural Networks (RNNs)~\cite{hochreiter1997long} instead of Multi-Layer Perceptrons (MLPs), as control policy networks.
For more complicated tasks with high-dimensional observation spaces, one can use Transformers~\cite{NIPS2017_3f5ee243} for better representation ability; for relatively simpler tasks, one can choose RNNs such as Long Short-Term Memory (LSTM)~\cite{hochreiter1997long} for its simplicity.
In this work, we use LSTM as our policy backbone, where the input is the history observations $\sigma(\hat{h}_t)$ and the output is action $a_{t}$.

%% file: simulation.tex
\vspace{-11pt}
\section{Experiments}
In our experiments, our main objective is to answer the following three questions.
\begin{itemize}[topsep=0ex,itemsep=-0.5ex]
    \item \textbf{Convergence}: can we learn an almost unexploitable policy profile using our approach in STLGames?
    In other words, can we approximately obtain the Nash joint policy profile?    \item \textbf{Efficiency}: is the policy convergence efficient in continuous environments for STLGame?
    \item \textbf{Robustness}: is an ego agent with the converged policy robust to unseen opponents?
\end{itemize}
Our approach is tested in two simulation benchmarks: Ackermann steering vehicles and autonomous drones.
We implement and compare both reinforcement learning and the STL gradient-based method as the best response policy synthesis approaches.
In our experiments, we focus on three metrics. First, the \textbf{exploitability} (\ref{eq:exploitability}) of the policy profiles found. Exploitability signals how close we are to the Nash equilibrium. Second, the STL \textbf{robustness value}. Robustness signals how much the control policy (un)satisfies the STL specification. Lastly, the \textbf{satisfication rate} of STL specification in rollouts. STL specifications are satisfied when the robustness value is non-negative.
\vspace{-7mm}
\begin{figure}[H]
    \centering
    \includegraphics[width=0.9\linewidth]{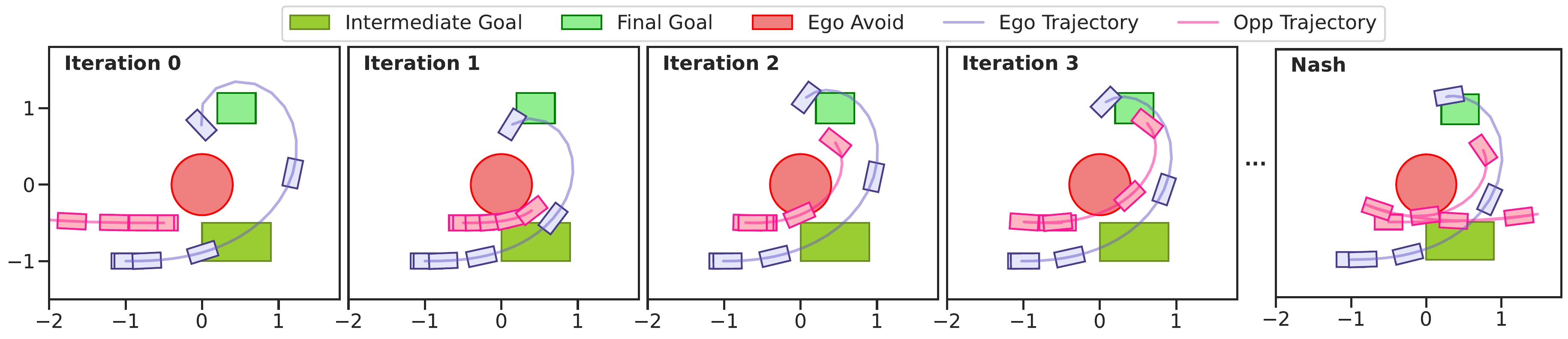}
    \vspace{-2mm}\caption{Vehicles trajectories sampled from some FSP iterations and finally from Nash profile.
    In the trajectories sampled from Nash profile, the opponent is trying its best to block the ego vehicle.}
    \label{fig:traj_through_iter_car}
\end{figure}
\vspace{-4mm}
\noindent \textbf{Ackermann Steering Vehicles.} The Ackermann steering vehicles are modeled by the following kinematic single-track dynamics. 
We consider two vehicles with the same dynamics in the environment, as shown in Figure~(\ref{fig:traj_through_iter_car}).
The state space model consists of 
$\bold{x} =
[s_x, s_y, \delta, v, \Psi]$, where $s_x$ and $s_y$ denote the position in $x$ and $y$ coordinate, $\delta$ is the steering angle, $v$ is the velocity, $\Psi$ is the heading angle, and $l_{wb}$ is the wheelbase. 
The control inputs are
$
[u_1, u_2] = [v_{\delta}, a_{long}]
$ where $v_{\delta}$ is the steering angle velocity and $a_{long}$ is the longitude acceleration. The continuous-time system is:
\begin{equation}
\begin{aligned}
\dot{x}_1= x_4 \cos \left(x_5\right),
\dot{x}_2= x_4 \sin \left(x_5\right),
\dot{x}_3= f_{steer}\left(x_3, u_1\right),
\dot{x}_4= f_{a c c}\left(x_4, u_2\right),
\dot{x}_5=\frac{x_4}{l_{wb}}\text{tan}(x_3).\nonumber
\end{aligned}
\end{equation}
where $f_{steer}$ and $f_{acc}$ impose physical constraints on steering and acceleration.
The system is discretized with sampling time $\Delta t=0.1s$.
The STL task is formally defined as (where $T=50$):
\begin{align}
    \phi = &\Diamond_{[0, T-1]}(q^{1}\in \text{Intermediate Goal}) \land \Diamond_{[0, T-1]}(q^{1}\in \text{Final Goal})\nonumber \\
    &\land \Box_{[0, T-1]}\neg(q^{1}\in \text{Red Circle}) \land ( \|q^1 - q^{-1}\|_2^2 \geq d_{\text{min}}^2).
\end{align}
Intuitively, the ego vehicle should eventually reach the intermediate goal and final goal, and always avoid the red dangerous region, and maintain a safe distance with the opponent vehicle.
The ego vehicle aims to satisfy $\phi$ and the opponent vehicle aims to satisfy $\neg \phi$.

\begin{figure}[H]
    \centering
    \includegraphics[width=0.275\linewidth]{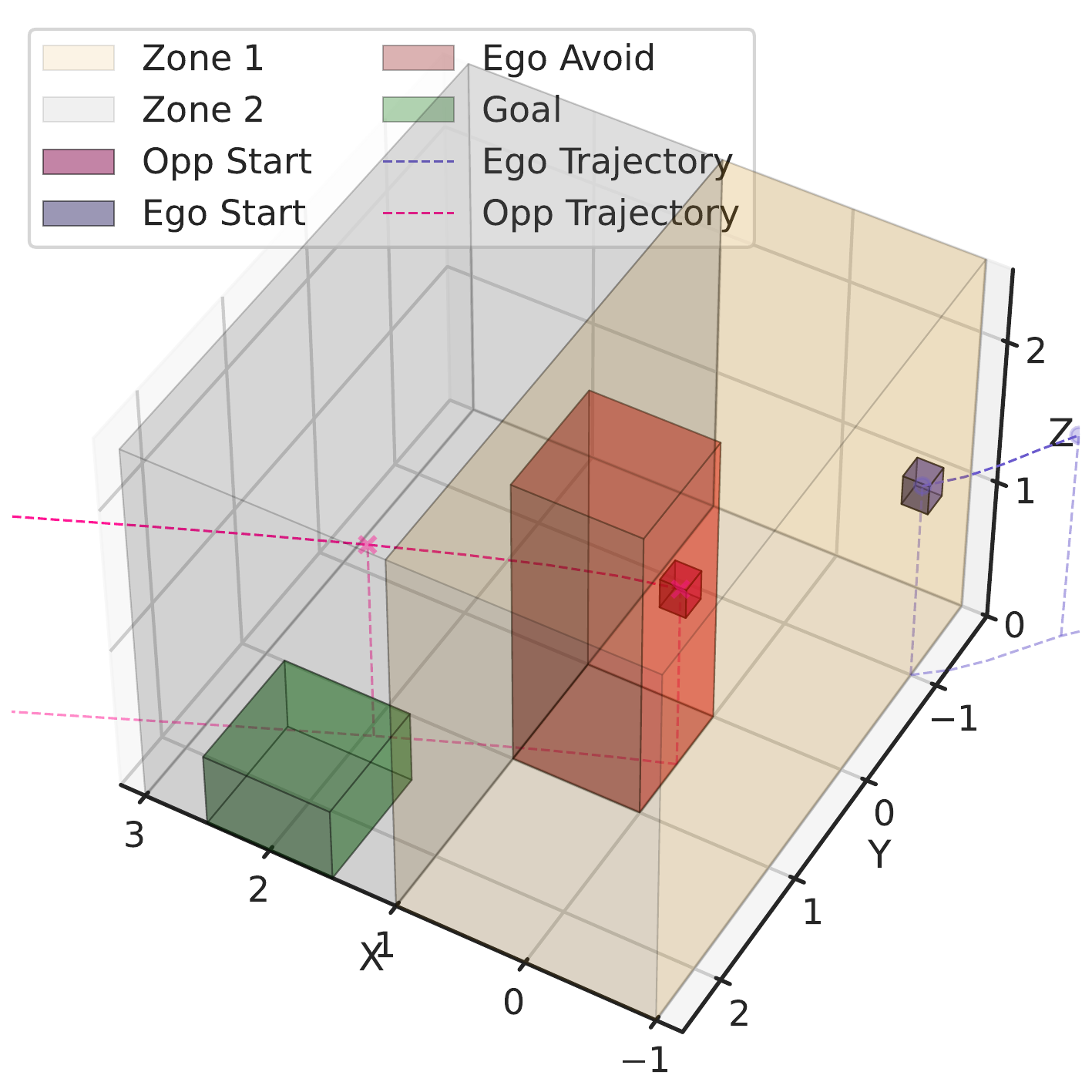}
    \includegraphics[width=0.275\linewidth]{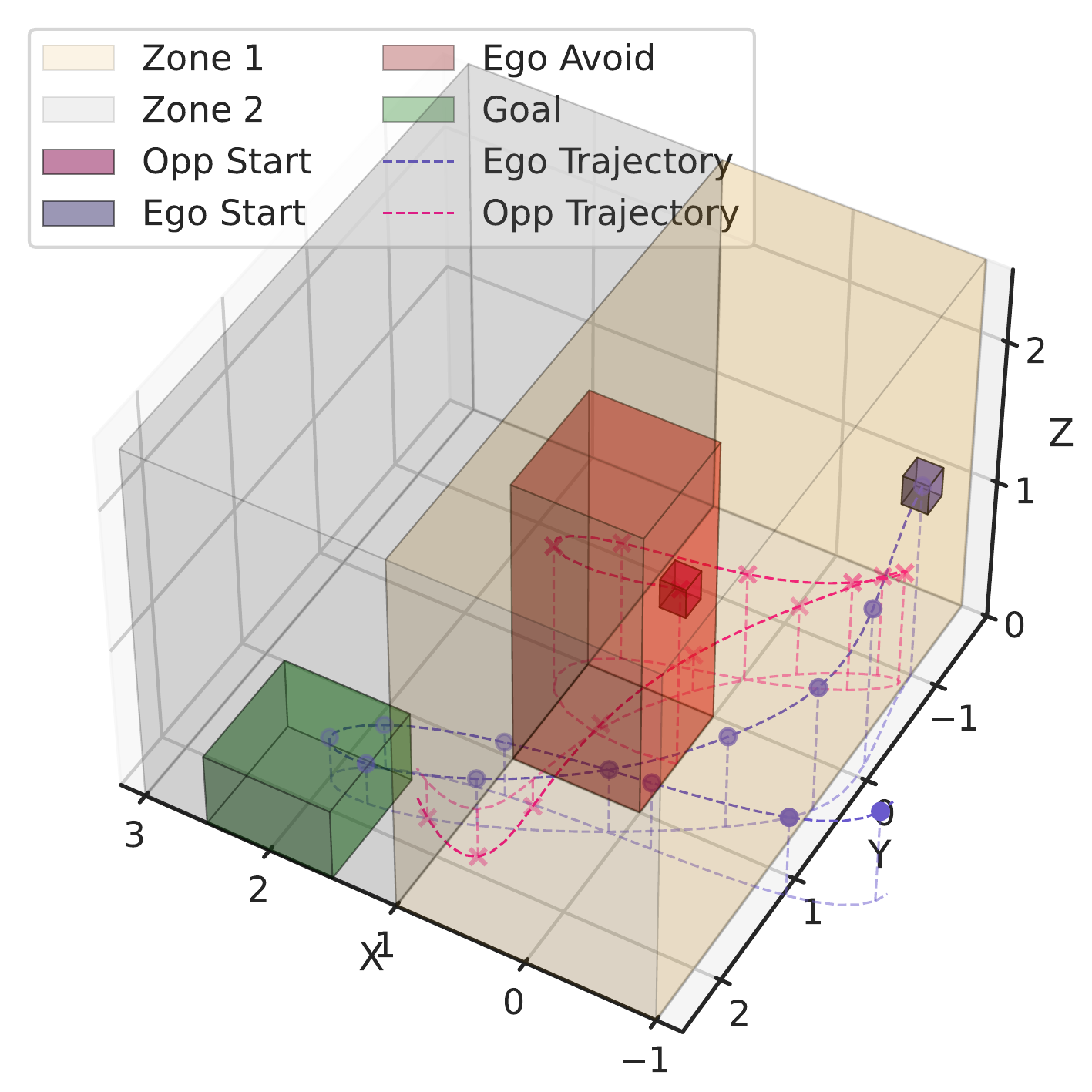}
    \includegraphics[width=0.275\linewidth]{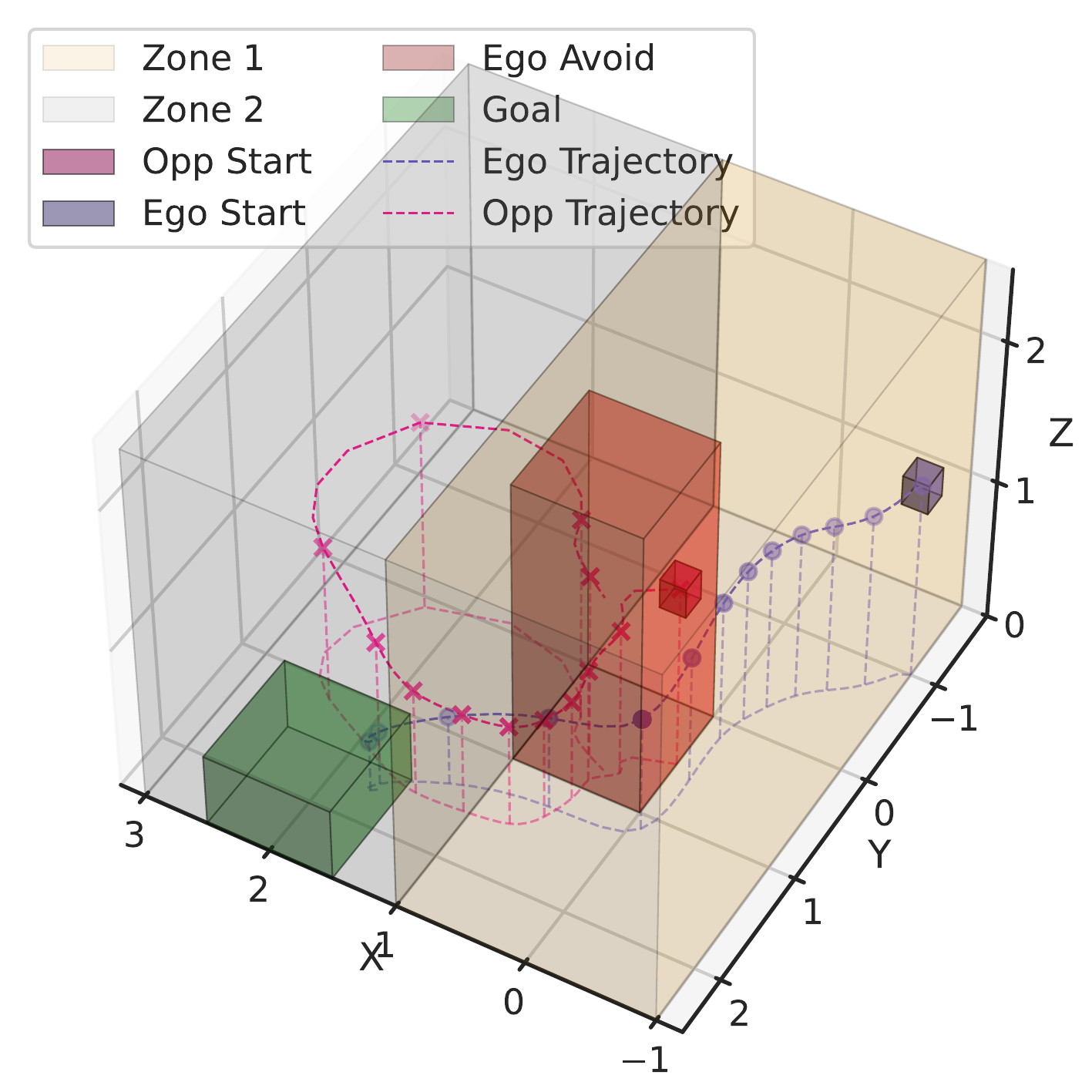}
    \vspace{-3mm}\caption{Drone trajectories from randomly initialized policy (Left) to FSP iteration1 (Middle) and finally Nash profile (Right). For better visualizations, please see our \href{https://sites.google.com/view/stlgame}{website}.}
    \label{fig:drone_nash}
\end{figure}
\begin{figure}[H]
    \centering
    \includegraphics[width=0.8\linewidth]{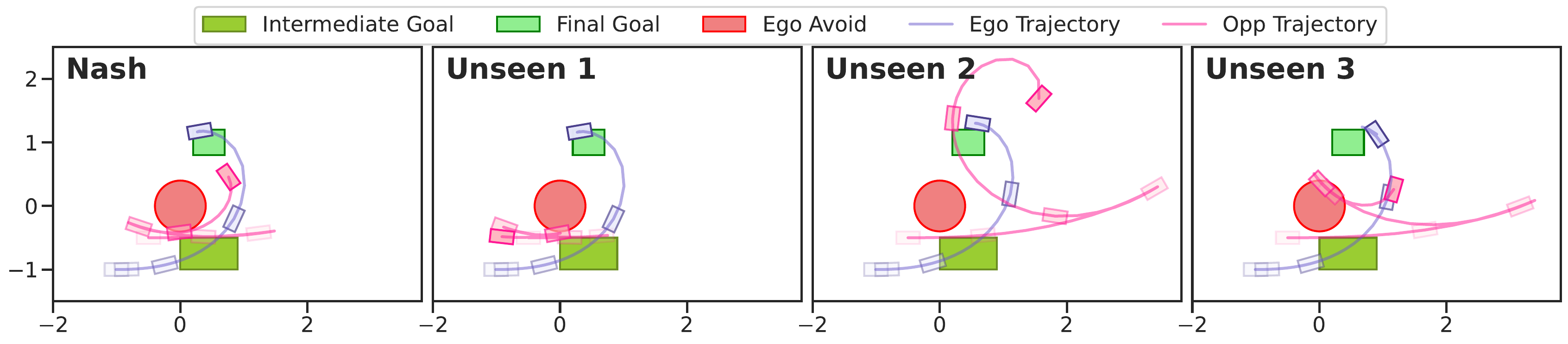}
    \caption{Sampled vehicles trajectories when the ego vehicle is playing a Nash policy, and the opponent is playing a Nash (Left), or some unseen policy (Unseen 1, Unseen 2, Unseen 3).}
\label{fig:nash_vs_unseen}
\end{figure}

\noindent\textbf{Autonomous Drones.}
Autonomous drones are nowadays widely deployed in various domains such as forestry, industry, and agriculture.
These deployed drones may face some uncertainty or even other unknown agents as adversaries in the environment.
Thus, it is critical to develop a robust control policy in the worst-case scenarios.
This is formulated as a zero-sum game between two drone players.
Specifically, we consider the 3D environment showed in Figure~(\ref{fig:drone_nash}), where the ego drone should 1) eventually arrive at the goal area, 2) always avoid the unsafe column, 3) always maintain safe distance with the opponent drone, and 4) always obey the altitude rules in different Zones.
They are formalized by the STL formula (similar to~\cite{pant2017smooth}):
\begin{align}
    \phi &= \Diamond_{[0, T-1]}(q^1 \in \text{Goal})\land\Box_{[0, T-1]} \big( \neg(q^1 \in \text{Unsafe}) \land ( \|q^1 - q^{-1}\|_2^2 \geq d_{\text{min}}^2) \big) \nonumber\\
&\quad \Box_{[0, T-1]} \big( q^1 \in \text{Zone}_1 \implies z^1 \in [1, 5]\big) \land\Box_{[0, T-1]} \big( q^1 \in \text{Zone}_2 \implies z^1 \in [0, 3]\big),
\end{align}
where $T=50$, and $q^1$ is the position of ego drone in the $x, y, z$ coordinate, and $q^{-1}$ is the position of opponent drone.
We use the same drone dynamics from~\cite{pant2015co, pant2017smooth} which have been shown successful in real-time quad-rotors control. 
\vspace{-3mm}
\vspace{-5mm}
\noindent \textbf{Exploitability Convergence and Efficiency.}
Exploitability measures how much worse the joint policy profile is compared to the Nash profile.
A policy is said to be less exploitable if it is closer to a Nash policy.
However, a challenge of computing exploitability (see its definition~(\ref{eq:exploitability})) is that it involves the computation of the best responses.
In this work, we empirically consider the best policy learned among all training epochs in the gradient-based method as the best response.
It takes around 10 minutes to train each FSP iteration on AMD Ryzen7 4800H.
\vspace{-3mm}
\begin{figure}[H]
    \centering    \includegraphics[width=0.34\linewidth]{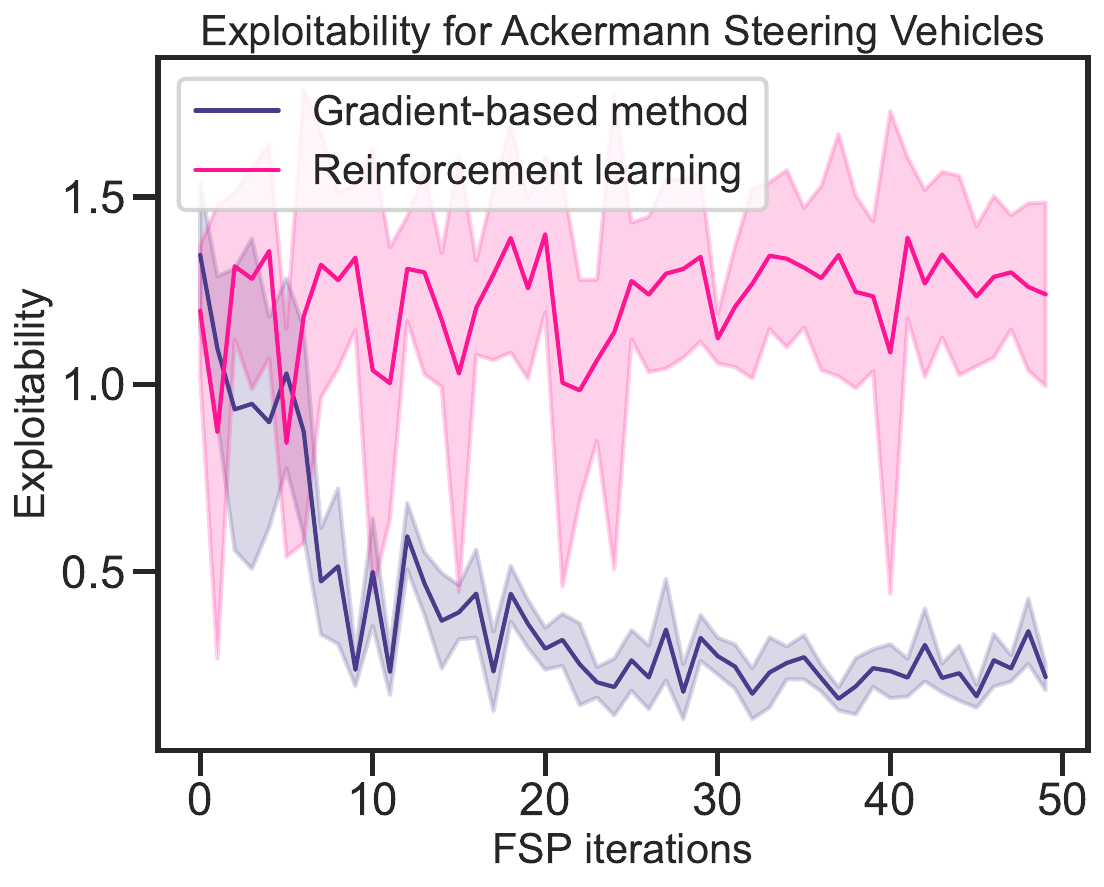}
    \includegraphics[width=0.33\linewidth]{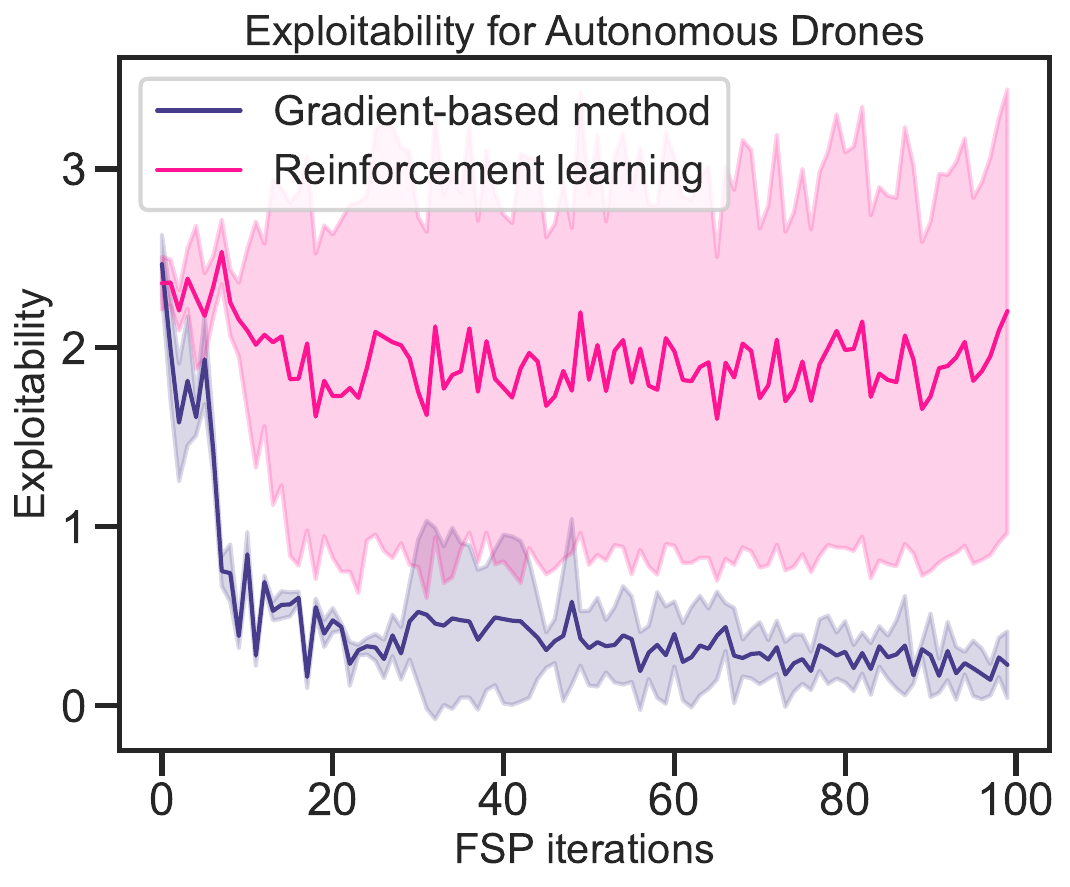}
    \vspace{-3mm}\caption{Exploitability for Ackermann Steering Vehicles (Left) and Autonomous Drones (Right). We run five different initial conditions for each environment.}
    \label{fig:exploitability}
\end{figure}
\vspace{-3mm}
The exploitability curves are plotted in Figure~(\ref{fig:exploitability}) for both benchmarks.
Both plots are run over five random initial conditions, and the solid lines denote the mean value, and the shaded region denotes the standard deviation.
We observe that FSP with gradient-based BR converges to almost zero exploitability, signaling that the final average policy profile is approximately Nash.
For RL-based BR, we use PPO with LSTM-based actor and critic under the same simulation budget, which fails to converge.
PPO struggled to learn effective response policies at each iteration due to sparse reward signals.
Interested readers can refer to our \href{https://sites.google.com/view/stlgame}{website} for more experimental results. 
\vspace{-2mm}
\begin{table}[H]
\centering
\begin{tabular}{ll|cc|cc}
\toprule
& & \multicolumn{2}{c|}{Seen Opponents} & \multicolumn{2}{c}{Unseen Opponents}\\
Scenario &Policy & STL Robustness & Satisfaction Rate & STL Robustness & Satisfaction Rate\\
\midrule
\multirow{2}{2em}{Vehicles} 
& Nash & $-0.06\pm 0.12$ & 18.7\% & $-0.12\pm 0.17$ & 16.7\% \\
& BR & $-0.02\pm 0.02$ & 20.0\% & $-0.21\pm 0.24$ & 2.0\% \\
\midrule
\multirow{2}{4em}{Drones} 
& Nash & $0.10\pm 0.06$ & 94.7\% & $0.01\pm 0.27$ & 88.0\% \\
& BR & $0.09\pm 0.05$ & 96.0\% & $-0.73\pm 0.16$ & 0.0\% \\
\bottomrule
\end{tabular}
\vspace{-2mm}
\caption{We play Nash policy and best response policy against seen policies and unseen policies. The best response is trained only against the average of seen policies.}
\label{table:robust}
\end{table}
\vspace{-4mm}
\noindent\textbf{Robustness against Unseen Opponents.}
Nash policy is the best choice for the ego agent, in the sense of worst-case return, when the ego agent has no prior information on the opponents' policy.
To test this, we collect elite opponent policies and split them into \emph{seen policy set} and \emph{unseen policy set}.
We train a best response policy to only the average of all seen policies.
The results are reported in Table~(\ref{table:robust}).
We can see that Nash policy is a more conservative in the sense that it is worse than the BR policy when playing against seen policies.
However, when it comes to unseen opponents, the BR policy performs significantly worse in terms of both robustness value and STL satisfaction rate.
This signals that the Nash policy can be deployed safely to unseen opponents while previously learned BR policy may fail drastically (e.g., in drones example).
Some Ackermann vehicle trajectory demonstrations of Nash policy against unseen opponents can be found in Figure~(\ref{fig:nash_vs_unseen}), and those for the drone experiment can be found on our \href{https://sites.google.com/view/stlgame}{website}. 

%% file: conclude.tex
\vspace{-26pt}
\section{Conclusion}
In this paper, we formulate STLGame for continuous environments, where ego agents aim to maximize STL satisfaction while opponent agents aim to minimize it. We propose an STL gradient-based method to learn a best response policy w.r.t. the opponents' average policy, under the classic fictitious play framework, and we show that it can converge to an almost unexploitable policy (Nash policy) effectively and efficiently. 
The STL gradient-based method significantly outperforms RL for best responses policy learning.
The converged policy is empirically shown to be robust to unseen opponent policies, and achieve higher STL robustness values and satisfaction rate compared to the best response policy trained from seen opponents. 
In the future, we will consider more agents inside the ego and opponent teams and aim to scale the multi-agent control synthesis under STL tasks.

%% file: append.tex
\newpage

\section{Appendix}

\subsection{Policy network details}
For STL gradient-based method, we use an LSTM network with one recurrent layer to process the state/observation sequence and use a linear layer as the action output head.
Note that in most control cases, MLPs are basically expressive enough to represent a performant control policy, for example, even in challenging autonomous vehicles racing~\cite{sun2023benchmark, sun2023mega}.
However, if the policy input is a long sequence such as all history observations, a powerful sequence model (RNNs or Transformers~\cite{NIPS2017_3f5ee243} is necessary to train a performant control policy.
For reinforcement learning-based best response policy synthesis, we use PPO~\cite{schulman2017proximal} and also use a shared LSTM network to embed sequence inputs, and two linear heads for actor and critic respectively.
The LSTM has the same architecture as the STL gradient-based best response policy.
For more hyper-parameters, please refer to our codebase~\url{https://github.com/shuoyang2000/STLgame}

\subsection{Reinforcement learning for best responses}
\begin{figure}[H]
    \centering
    
    \includegraphics[width=0.49\linewidth]{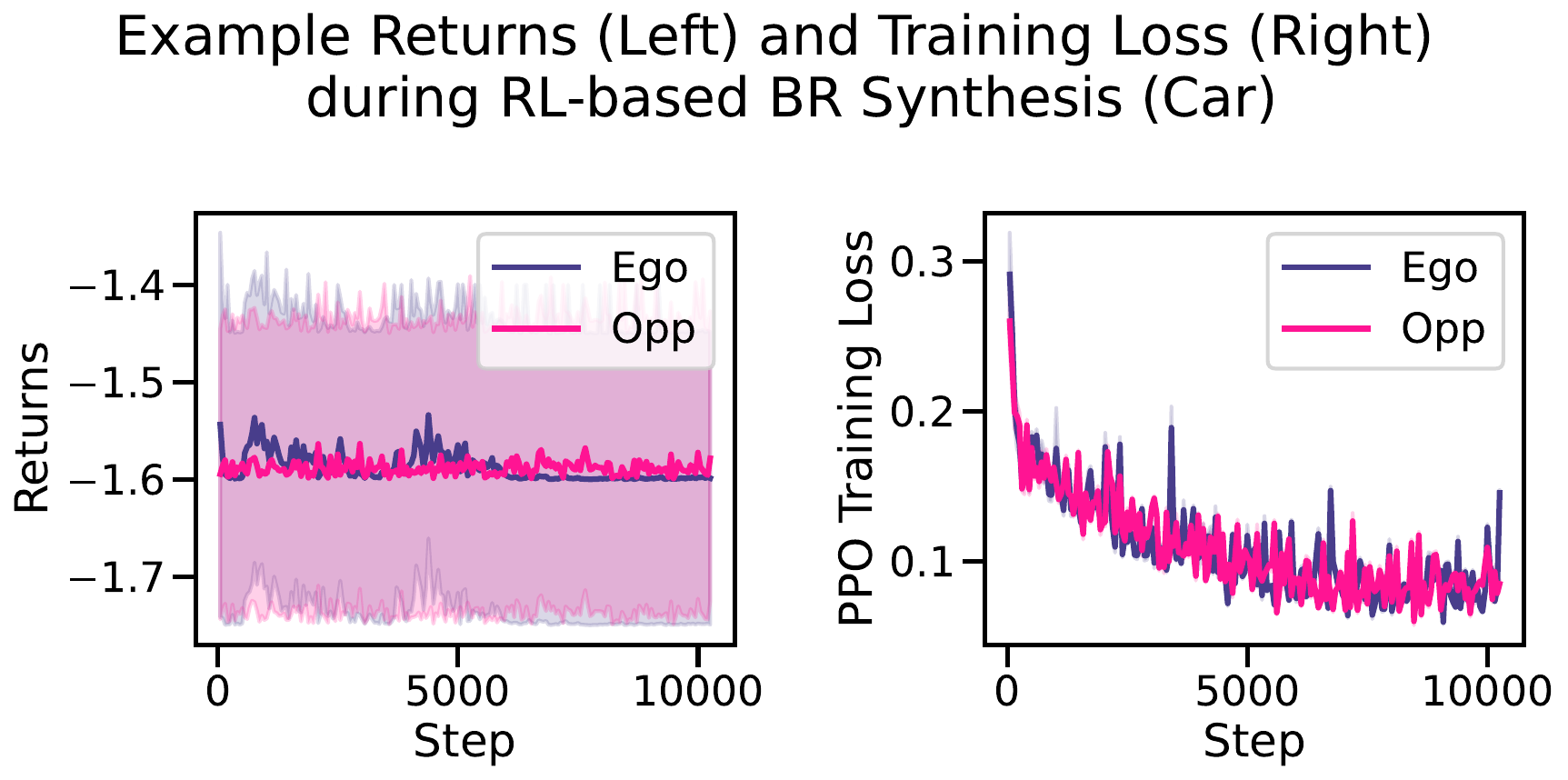}
    \includegraphics[width=0.49\linewidth]{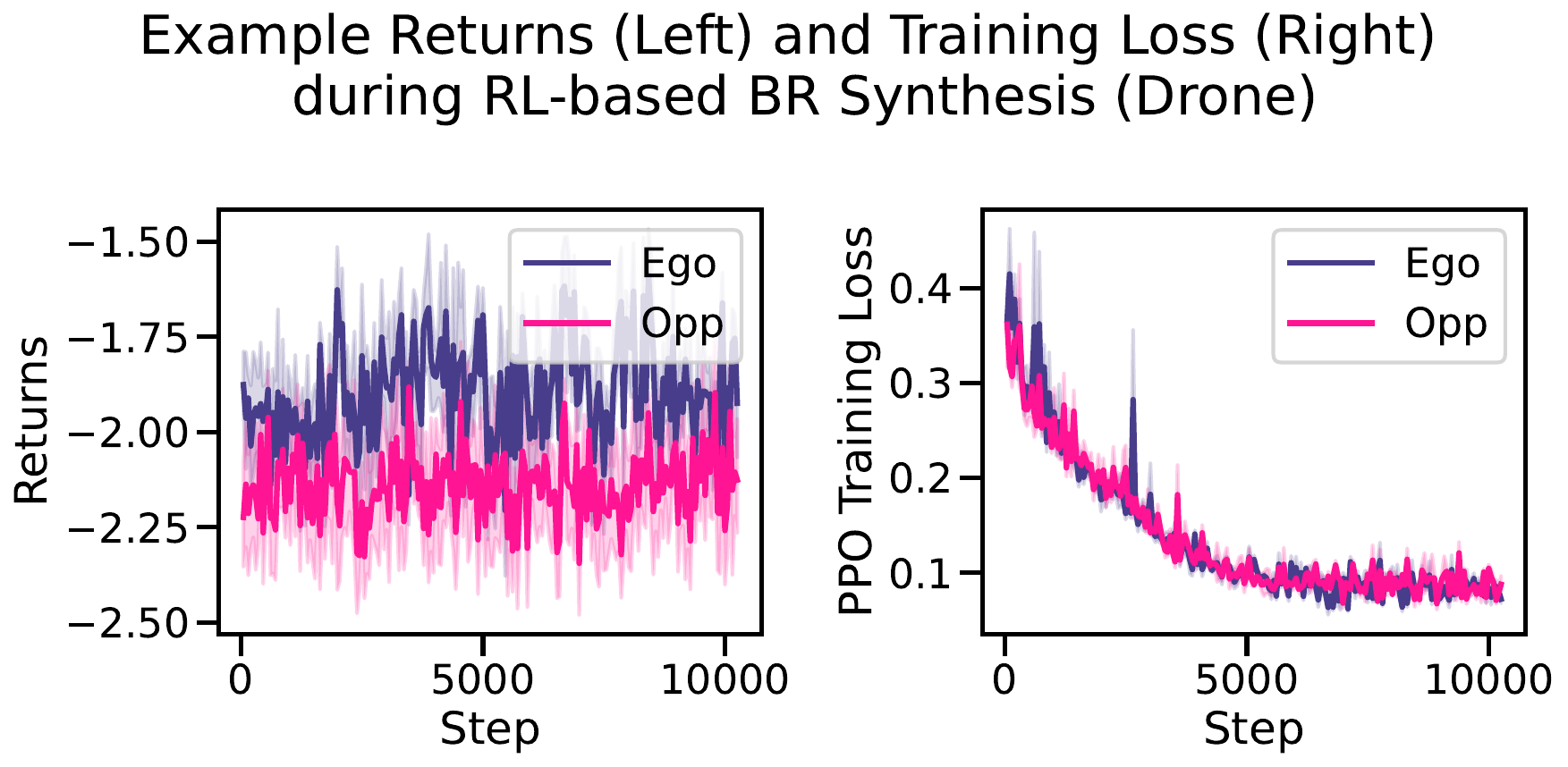}

    \caption{PPO training return curves and loss curves. 
    Ego agent is expected to maximize the return and opponent agent is expected to minimize the return. 
    All results are run on 5 different initial conditions.
    Left: Ackermann steering vehicles; right: autonomous drones.}
    \label{fig:ppo_return_loss}
\end{figure}

PPO is run with the same computation budget, i.e., the overall training simulation steps is the same as the STL gradient-based method, so the overall training time is also similar.
The required simulation steps per FSP iteration is:
\begin{align}
    \text{training epochs} \times \text{opponent samples} \times \text{trajectory horizon} = 200\times 15\times 50=1.5\times 10^5. \nonumber
\end{align}
The exploitability convergence comparison can be found in Figure~(\ref{fig:exploitability}).
We also include the training return curves and loss curves in Figure~(\ref{fig:ppo_return_loss}).
We can observe that the loss indeed has been minimized but the return is not improved.
A key reason is that the STL robustness-based reward function~(\ref{eq:reward_func_robustness}) is sparse, and also model-free RL training is not very sample-efficient and the learned response policy is thus far away from the best response policies.
We are currently granting RL training more computation budget to see if RL-based BR can facilitate FSP to converge to Nash equilibrium in STLGame.
New experimental results will be uploaded.

\subsection{Autonomous drones dynamical systems}

We use the same drone dynamics from~\cite{pant2015co, pant2017smooth} which have been shown successful in real-time quad-rotors control. 
The final discretized quad-rotor system is modeled as follows:
\begin{align}
\left[\begin{array}{c}\dot{x}_{k+1} \\ \dot{y}_{k+1} \\ \dot{z}_{k+1} \\ x_{k+1} \\ y_{k+1} \\ z_{k+1}\end{array}\right]
    =
\left[\begin{array}{cccccc}1 & 0 & 0 & 0 & 0 & 0 \\ 0 & 1 & 0 & 0 & 0 & 0 \\ 0 & 0 & 1 & 0 & 0 & 0 \\ 0.2 & 0 & 0 & 1 & 0 & 0\\ 0 & 0.2 & 0 & 0 & 1 & 0\\ 0 & 0 & 0.2 & 0 & 0 & 1\end{array}\right]
 \left[\begin{array}{c}\dot{x}_{k} \\ \dot{y}_{k} \\ \dot{z}_{k} \\ x_{k} \\ y_{k} \\ z_{k}\end{array}\right]
    +
\left[\begin{array}{ccc}1.96 & 0 & 0 \\ 0 & -1.96 & 0 \\ 0 & 0 & 0.4 \\ 0.196 & 0 & 0\\ 0 & -0.196 & 0 \\ 0 & 0 &0.04\end{array}\right]
\left[\begin{array}{c}
    \theta_k \\ \varphi_k \\\tau_k\end{array}\right],\nonumber
\end{align}
In particular, the system state is composed of the position and velocity in the $x, y, z$ coordinates, and the control input includes
the desired roll angle, pitch angle, and thrust.

\subsection{Ego and opponent agents initial conditions}
Note that the initial conditions affect the final STL robustness values and satisfaction rates a lot.
If the opponent agents start from a clearly advantageous position, then even if the ego agent deploys a Nash policy, the STL satisfaction rate may still be low.
Consider a simple example, if the opponent agents' initial position is exactly the ego agents' destination, then it is possible that ego agents can never reach the destination and avoid crash with opponents simultaneously no matter which policy is adopted, implying that the STL specification can never be satisfied.
Note that this also signals that the final absolute values (e.g., STL robustness and satisfaction rates) of Nash policy is not so important, because it might be affected by many factors, such as initial conditions, environment setup, STL specification, and ego and opponent agents dynamics.
The relative values are more important because they can demonstrate that Nash policy is the best choice among all policies when facing an a-priori unknown opponent.
We try to consider various initial condition setups in our experiments and avoid those initial conditions showing clear advantage for any side.
For Ackermann steering vehicles case:
\begin{align}
    \left[\begin{array}{c}
    x_{ego} \\ y_{ego}\end{array}\right]&\in \left\{
    \left[
    \begin{array}{c}
    -1 \\ -1\end{array} \right],
    \left[
    \begin{array}{c}
    -0.5 \\ -1\end{array} \right],
    \left[
    \begin{array}{c}
    -1 \\ -0.5\end{array} \right],
    \left[
    \begin{array}{c}
    -0.75 \\ -0.75\end{array} \right],
    \left[
    \begin{array}{c}
    -0.5 \\ -0.75\end{array} \right]
    \right\},\nonumber\\
    \left[\begin{array}{c}
    x_{opp}\nonumber \\ y_{opp}\end{array}\right]&\in \left\{
    \left[
    \begin{array}{c}
    -0.5 \\ -0.5\end{array} \right],
    \left[
    \begin{array}{c}
    0 \\ 0\end{array} \right],
    \left[
    \begin{array}{c}
    -0.25 \\ -0.25\end{array} \right],
    \left[
    \begin{array}{c}
    -0.5 \\ -1\end{array} \right],
    \left[
    \begin{array}{c}
    -0.5 \\ -0.9\end{array} \right]
    \right\}.\nonumber
\end{align}
For autonomous drones case:
\begin{align}
    \left[\begin{array}{c}
    x_{ego} \\ y_{ego}\\ z_{ego}\end{array}\right]&\in \left\{
    \left[
    \begin{array}{c}
    -1 \\ -1\\ 1.4\end{array} \right],
    \left[
    \begin{array}{c}
    -0.5 \\ -1\\ 1.1\end{array} \right],
    \left[
    \begin{array}{c}
    -1 \\ -0.5\\ 1.5\end{array} \right],
    \left[
    \begin{array}{c}
    0.5 \\ -0.75\\ 1.2\end{array} \right],
    \left[
    \begin{array}{c}
    0 \\ -0.75\\ 1.2\end{array} \right]
    \right\},\nonumber\\
    \left[\begin{array}{c}
    x_{opp}\nonumber \\ y_{opp}\\ z_{opp}\end{array}\right]&\in \left\{
    \left[
    \begin{array}{c}
    0 \\ 0.5\\ 1.3\end{array} \right],
    \left[
    \begin{array}{c}
    0 \\ 0\\ 1.1\end{array} \right],
    \left[
    \begin{array}{c}
    -0.25 \\ -0.25\\ 0.8\end{array} \right],
    \left[
    \begin{array}{c}
    -0.5 \\ -1\\ 0.8\end{array} \right],
    \left[
    \begin{array}{c}
    -0.5 \\ -0.9\\ 1.4\end{array} \right]
    \right\}.\nonumber
\end{align}
All other state variables such as velocity or angles are initialized by 0.